\documentclass[letterpaper,journal]{IEEEtran}
\usepackage{amsmath,amsfonts}
\usepackage{algorithmic}
\usepackage{algorithm}
\usepackage{array}
\usepackage[caption=false,font=normalsize,labelfont=sf,textfont=sf]{subfig}
\usepackage{textcomp}
\usepackage{stfloats}
\usepackage{url}
\usepackage{verbatim}
\usepackage{graphicx}
\usepackage{cite}
\usepackage{booktabs}
\usepackage{graphicx}
\usepackage{capt-of}
\usepackage{comment}
\usepackage{pifont}
\usepackage{booktabs}
\usepackage{makecell}

\urldef{\projecturl}\url{https://anonymous.4open.science/r/BeyondContact-Anonymized/}

\begin{document}
\setlength{\abovedisplayskip}{2pt}
\setlength{\belowdisplayskip}{2pt}
\setlength{\abovedisplayshortskip}{0pt}
\setlength{\belowdisplayshortskip}{2pt}
\setlength{\jot}{1pt}

\title{BeyondContact: A Reinforcement Learning Framework for Airflow-Based Non-Contact Manipulation}

\author{Peiyu Yang$^{*}$, Mohamad S. Moudallal$^{*}$, Antonio Langella, Giulio Turrisi, \\Victor Barasuol, and Claudio Semini
\thanks{All the authors are with the Dynamic Legged Systems lab, Istituto Italiano di Tecnologia (IIT), 16163, Genova, Italy. E-mail: name.surname@iit.it. }
\thanks{$^{*}$ Peiyu Yang and Mohamad S. Moudallal contributed equally.}%
}

\IEEEaftertitletext{%
    \vspace{-10mm}
    \begin{minipage}{\textwidth}
        \centering
        \includegraphics[width=0.95\linewidth]{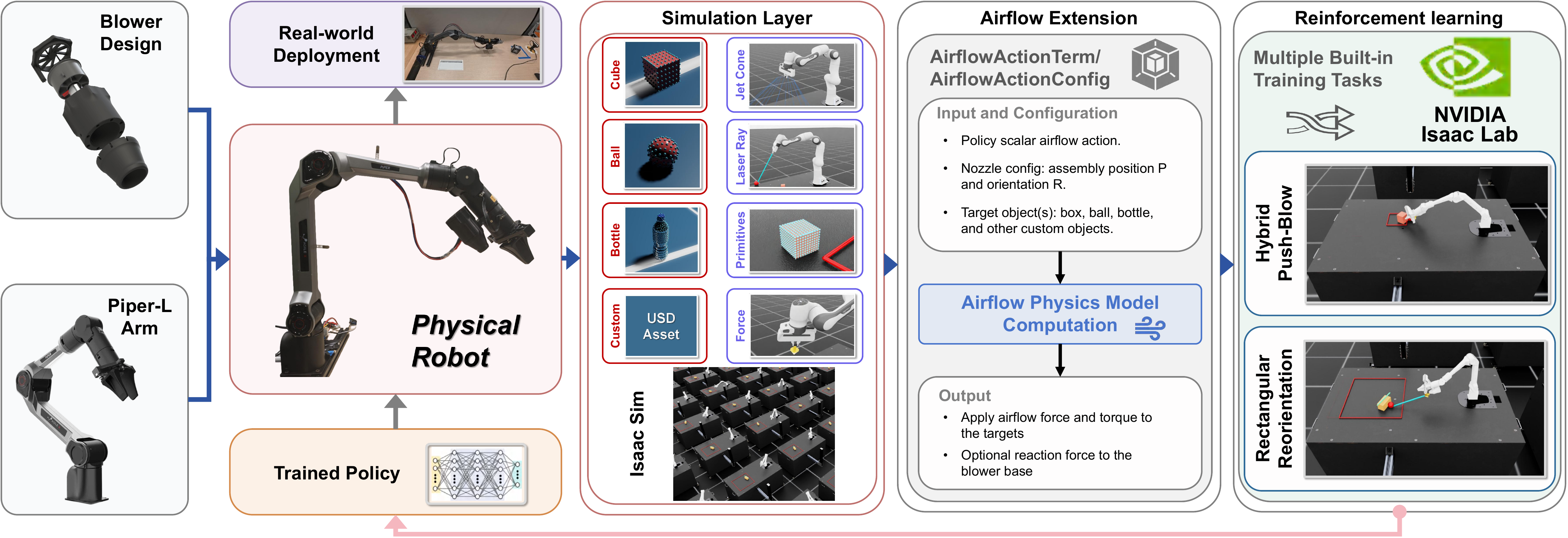}
        \vspace{-1mm}
        \captionof{figure}{Overview of the proposed reinforcement learning (RL) framework for airflow-based non-contact manipulation. A custom blower is integrated with an AgileX Piper-L robotic arm. The airflow simulation extension supports multiple object types and nozzle configurations, computing and applying airflow-induced forces and torques in simulation. The extension enables RL across multiple tasks, including hybrid push–blow (HPB) and rectangular reorientation (RR), with the trained policies subsequently deployed on the physical robot. 
        }
        \label{Fig:Framework-overview}
        \vspace{0.5\baselineskip}
    \end{minipage}
}

\maketitle

\begin{abstract}
    Contact-based robotic manipulation is constrained by end-effector reachability and the mechanical properties of objects. Airflow can exert forces without direct robot-object contact, enabling manipulation of objects beyond the reach or when direct contact is undesirable. However, the lack of suitable simulation has hindered the application of reinforcement learning to airflow-based manipulation, limiting the broader adoption of this mechanism compared with contact-based methods. To address this gap, we developed a robot- and task-agnostic airflow extension for Isaac Lab that supports massively parallel training. The extension uses experimentally identified airflow models to compute the forces and torques exerted on objects. Based on this extension, we train a hybrid push–blow (HPB) policy that combines contact-based pushing with airflow manipulation and a rectangular reorientation (RR) policy that controls a cuboid's heading using airflow alone. To enable physical execution, we design a lightweight, high-speed blower that coexists with a gripper. In zero-shot real-world deployment, HPB achieves an 80\% success rate over 20 trials, while RR achieves a mean absolute heading error of 2.6\(^\circ\) across four target headings. The project materials are available at \textbf{https://github.com/iit-DLSLab/BeyondContact}.
\end{abstract}
\maketitle


\vspace{-2mm}
\section{Introduction}\label{Sec:Introduction}

\IEEEPARstart{R}{ecent} advances in robotic arm hardware and control methodologies have broadened the application of robotic manipulation in industrial\cite{rossini2026concert} and domestic settings\cite{spahn2024demonstrating}. However, prevailing manipulation paradigms remain predominantly based on direct robot-object contact, such as grasping\cite{turrisi2026binwalker} and pushing\cite{ren2025object}. Although contact-based manipulation is effective in conventional tasks, including assembly, pick-and-place, and object transport, its applicability is constrained by the robot's kinematic workspace, collision-free reachability, and the mechanical properties and geometry of the object. In contrast, non-contact manipulation enables robots to act on objects without direct robot-object contact, offering the potential for safer operation and remote manipulation\cite{al2022non}. Among other already exploited non-contact manipulation means (e.g., acoustic \cite{bai2026robotic}, magnetic \cite{stewart2025non}, and optoelectronic \cite{jia2025non}), airflow is particularly attractive because it can be generated using compact hardware and can extend the action range beyond the reach of conventional end-effectors while acting on multiple objects simultaneously.

Airflow actuation mechanisms have demonstrated the feasibility of non-contact object manipulation. Hyungpil \emph{et al.} \cite{moon2006distributed} proposed planar manipulation using two airflow sinks; Artur \emph{et al.} \cite{kopitca20263d} achieved three-dimensional manipulation of lightweight objects using a \(3\times3\) array of computer fans; and Bao \emph{et al.} \cite{bao2026development} developed a Bernoulli-based non-contact manipulator, in which compressed air is directed toward the surface to create a low-pressure region and hold the object. 

Other studies have explored directed airflow from blowers and compressed-air nozzles for manipulation tasks. Lee \emph{et al.}\cite{lee2022development} and Natsu \emph{et al.}\cite{natsu2023development} mounted blowers on mobile platforms and used manual teleoperation and predefined trajectories for large-area leaf collection. The studies in \cite{haeri2024meter} and \cite{kopitca2025remote} developed a compressed-air system with a two-degree-of-freedom (DoF) nozzle mounted at a fixed height to analyze the jet-induced airflow field over a planar surface. The former employed a heuristic proportional feedback controller to manipulate objects, whereas the latter combined identified object dynamics with a cross-entropy method (CEM)-based controller to perform path following, aggregation, and sorting of single and multiple objects. These studies demonstrate the potential of using directed airflow, but they relied on specially designed control strategies tailored to specific tasks.

Reinforcement learning (RL) offers a possibility to broaden the application by learning policies through interaction. Several studies have worked in this direction. Wu \emph{et al.}\cite{wu2022learning} investigated RL-based control of a mobile blower to collect and aggregate small objects. DextAIRity~\cite{xu2022dextairity} equipped one robotic arm with a commercial blower for deformable object manipulation. By coordinating grasping with two arms and blowing with the third, the system performed cloth unfolding and bag opening. However, policy training in these studies either relied on oversimplified virtual-particle models that do not explicitly capture the spatial decay and spread of the airflow velocity field or required extensive and time-consuming data collection. 

These limitations highlight the need for a computationally efficient airflow–object interaction module integrated into existing contact-based simulation platforms. Such a module, together with systematic parameter identification and experimental validation, would facilitate adaptation to diverse robotic platforms and custom blower configurations, policy training in simulation, and sim-to-real deployment.

To address this gap, we present an airflow-interaction extension for Isaac Lab\cite{nvidia_isaaclab}, which supports GPU-accelerated, high-throughput robot learning. The extension incorporates a configurable airflow-field model characterized by a centerline-decay function and a spatially varying radial profile, and maps local airflow velocities to forces and torques applied to simulated rigid bodies of various shapes. It can also be adapted to different robotic platforms and nozzle configurations while maintaining computational efficiency for parallel RL training. Building upon this extension, we present two RL training frameworks for the blower-equipped AgileX Piper-L robotic arm: hybrid push--blow (HPB), which coordinates contact pushing and non-contact airflow actuation through a unified policy to move objects to target regions beyond the arm's direct reach; and rectangular reorientation (RR), which uses airflow alone to align objects with target orientations. Finally, the learned policies are deployed and evaluated on a real robotic arm. The overall architecture is illustrated in Fig.~\ref{Fig:Framework-overview}.

To summarize, the main contributions of this work are:
\begin{itemize}
    
    \item The development of a configurable and computationally efficient airflow-interaction extension for Isaac Lab, supporting massively parallel robot-learning environments.
    \item The development and experimental validation of centerline-decay and radial-spreading models for custom blower and nozzle configurations, supported by a lightweight blower designed to coexist with the gripper.
    \item The design of two RL training frameworks that enable hybrid push--blow manipulation in HPB and purely airflow-driven reorientation control in RR, and the validation of the learned policies through zero-shot sim-to-real deployment.
\end{itemize}

The remainder of this paper is organized as follows: Section~\ref{Sec:Modeling} presents the airflow model. Section~\ref{Sec:Extension} introduces the simulation extension, while Section~\ref{Sec:Mechatronic} describes the mechatronic design. Section~\ref{Sec:Policy} describes the training frameworks. Section~\ref{Sec:Experiment} presents the related experimental setup and results, and Section~\ref{Sec:Conclusion} concludes the paper and discusses the future work.

\vspace{-3mm}

\section{Airflow Dynamics Modeling} 
\label{Sec:Modeling}

\begin{figure}[t]
    \centering
    \includegraphics[width=0.95\linewidth]{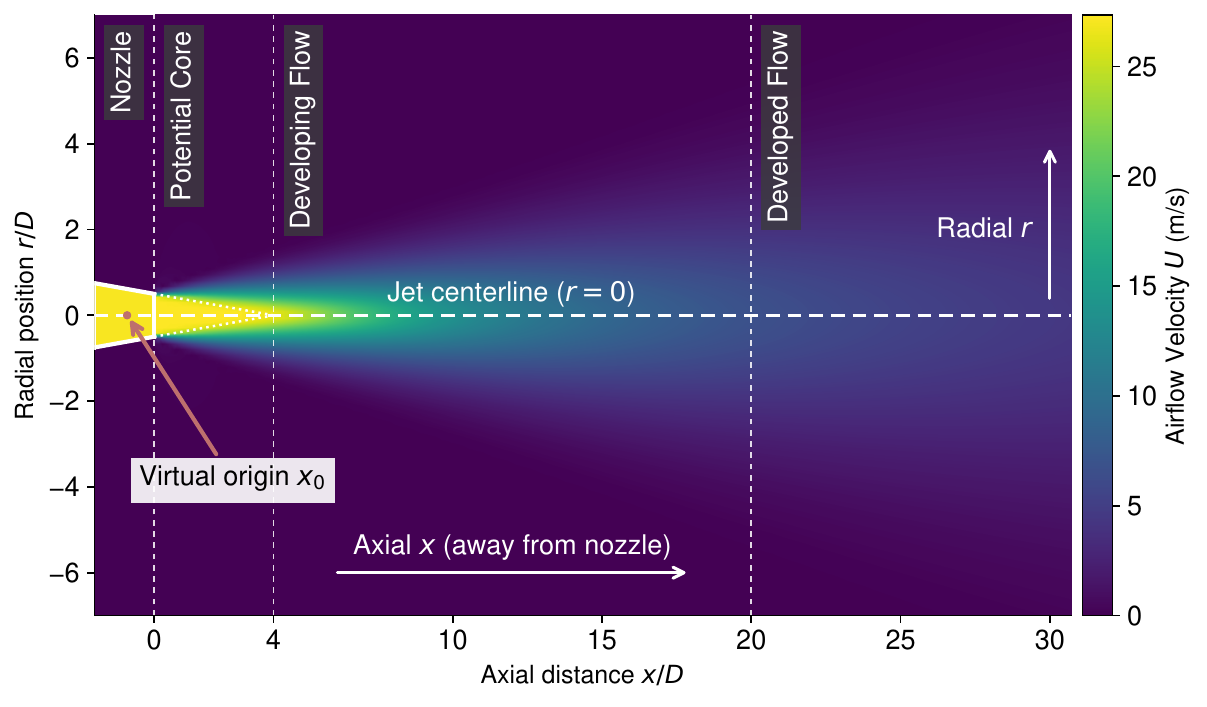}
    \vspace{-2mm}
    \caption{Simulated airflow velocity field of a free turbulent round jet. Axial and radial positions are normalized by the nozzle outlet diameter \(D\).}
    \label{fig:jet_model}
    \vspace{-5mm}
\end{figure}

This section introduces the simplified airflow velocity field models used to compute drag forces on objects in our Isaac Lab extension. We model the airflow generated by a blower as a time-averaged, quasi-steady free turbulent round jet \cite{albertson1950diffusion}. In practical terms, the model predicts the average airflow velocity field from a circular nozzle into otherwise still air, without resolving rapid turbulent fluctuations or the effect of nearby surfaces. The model describes how the jet decays and spreads with distance from the nozzle along two directions, axially along its centerline, and radially outward from it, as shown in Fig.~\ref{fig:jet_model}. In the following, we first define the axial and radial coordinates relative to the nozzle, then model how the velocity decays downstream along the centerline, and finally describe how it spreads radially.

\vspace{-2mm}
\subsection{Nozzle-Centered Representation}

Let $\mathbf{p}_0$ be the nozzle center and $\hat{\mathbf{e}}_n$ be the unit vector along the nozzle axis. The axial and radial coordinates of a spatial point $\mathbf{p}$ in the nozzle-centered coordinate system are defined as:
\begin{equation}
    x=(\mathbf{p}-\mathbf{p}_0)^\top\hat{\mathbf{e}}_n,
    \label{eq:axial_coordinate}
\end{equation}
\begin{equation}
    r = \left\|(\mathbf{p}-\mathbf{p}_0)-x\hat{\mathbf{e}}_n\right\|_2.
    \label{eq:radial_coordinate}
\end{equation}

The above model applies for $x>0$. Assuming a non-swirling turbulent free round jet, the mean transverse velocity components $U_y$ and $U_z$ in the nozzle coordinate system are negligible with respect to $U_x$~\cite{basset2022entrainment}, and the local airflow velocity vector is given by the following equation:
\begin{equation}
    \mathbf{u}_{\mathrm{air}}(\mathbf{p}) = U_x(x,r)\hat{\mathbf{e}}_n = U(x,r)\hat{\mathbf{e}}_n,
    \label{eq:airflow_vector_field}
\end{equation}
where $U(x,r)$ denotes the local airflow speed.

\vspace{-2mm}
\subsection{Centerline Jet Decay}

Existing turbulent round jet models define an initial \textit{potential core} region followed by a \textit{developing} region, and a far field \textit{developed} region where the axial flow gradient has achieved a near zero steady-state condition \cite{abramovich1984theory}, as visualized in Fig.~\ref{fig:jet_model}. The potential core is established by a region that starts at the nozzle outlet and ends approximately between $4D$ and $6D$, where \(D\) is the nozzle outlet diameter, and the centerline velocity remains identical or close to the outlet velocity $U_0$. Further downstream beyond the potential core, the airflow transitions into a self-similar jet, which is approximately modeled as an inverse relationship between axial velocity and centerline distance from the jet outlet.

Following this modeling approach, we formulate the centerline velocity as given in \eqref{eq:centerline_velocity}, where the flow is assumed to be incompressible and axially symmetric about its centerline \cite{hussein1994velocity, ball2012flow}:
\begin{equation}
    U_c(x) = U_0 \cdot f_{\mathrm{model}}(x),
\label{eq:centerline_velocity}
\end{equation}
where $f_{\mathrm{model}}(x)$ is the dimensionless decay function associated with a chosen airflow model. We consider two candidate models in this work: an exponential model and a shifted inverse model \cite{abramovich1984theory}.

The exponential model is defined as:
\begin{equation}
    f_{\mathrm{exp}}(x)=
    \begin{cases}
        1, & 0 < x \leq x_c, \\
        \exp[-k(x-x_c)], & x>x_c,
    \end{cases}
\label{eq:exp_model}
\end{equation}
where $x_c$ is the aforementioned potential core axial length and $k>0$ is the decay coefficient. This model provides a bounded representation and forms a compact empirical approximation within a finite operating range. Moreover, the model does not seek to replicate the asymptotic properties of a fully turbulent jet, because it tends to zero faster than an inverse power-law function.

The shifted inverse model is inspired by the Albertson three-zone model~\cite{albertson1950diffusion} and is defined as:
\begin{equation}
    f_{\mathrm{inv}}(x)=
    \begin{cases}
        1, & 0 < x \leq x_{\mathrm{inv}},\\[2pt]
        \dfrac{BD}{x-x_0}, & x > x_{\mathrm{inv}},
    \end{cases}
    \label{eq:inv_model}
\end{equation}
where $B$ is the dimensionless decay coefficient and $x_0$ is the virtual origin of the jet airflow cone, which does not necessarily coincide with the nozzle outlet $\mathbf{p}_0$. The transition point $x_{\mathrm{inv}}=x_0+BD$ makes the two branches continuous and keeps the predicted centerline velocity at $U_0$ near the nozzle.

\vspace{-3mm}
\subsection{Radial Jet Spreading}

Following the Albertson three-zone model, we model the radial decay of the mean velocity using a Gaussian profile:
\begin{equation}
    U(x,r) = U_c(x) \cdot
    \exp\left[
    -\frac{r^2}{2\sigma_r^2(x)}
    \right],
\label{eq:gaussian_model}
\end{equation}
where $\sigma_r(x)$ represents the spatial jet width across the radial dimension, and is modeled approximately as a linear spreading of the turbulent round jet:
\begin{equation}
    \sigma_r(x)=a_\sigma+Sx,
\label{eq:sigma_model}
\end{equation}
where $a_\sigma$ is the extrapolated width intercept and $S$ is the dimensionless spreading rate. The intercept accounts for the finite initial jet width and avoids forcing the spreading line through the nozzle origin.

Consequently, \eqref{eq:centerline_velocity}--\eqref{eq:sigma_model} describe axial decay and radial spreading separately. All the model parameters, including the nozzle exit velocity $U_0$, centerline-decay parameters, and radial-spreading parameters, are identified from experimental measurements of the blower with and without a nozzle, as presented in Sec.~\ref{Sec:Experiment}-A.

\vspace{-3mm}
\section{Airflow Extension} \label{Sec:Extension}

\begin{figure}[t]
    \centerline{
        \includegraphics[width=0.95\linewidth]{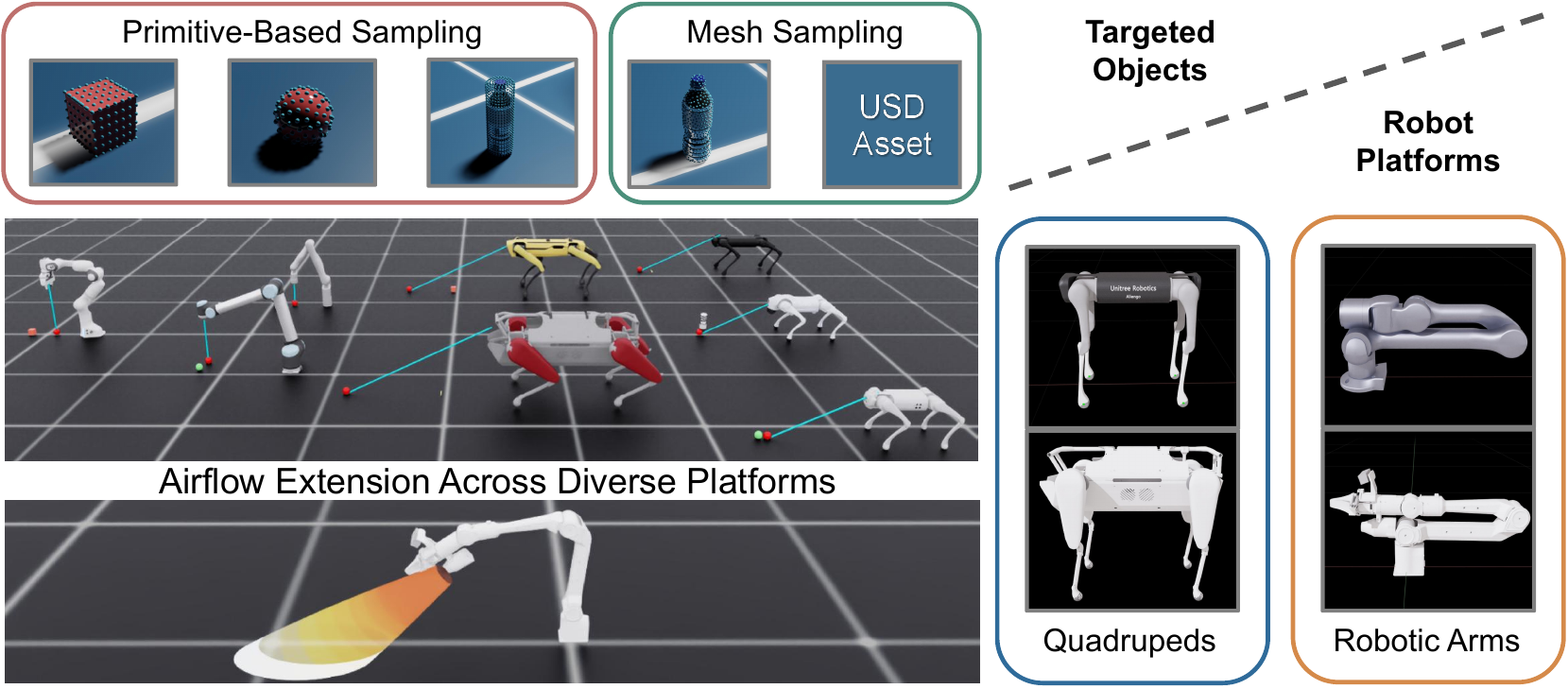}}
    \vspace{-2mm}
    \caption{Example of the airflow extension across diverse target objects and robotic platforms. Surface primitives are sampled from basic geometric shapes or mesh geometry in USD assets, while blowers can be mounted on various platforms, including quadrupeds and robotic arms.}
    \label{fig:showcase}
\end{figure}

This section presents the implementation of the airflow-interaction extension and its integration into Isaac Lab for large-scale robot learning. The extension maps the airflow dynamics model established in Sec.~\ref{Sec:Modeling} to forces and torques acting on target objects through batched PyTorch tensor operations. These forces and torques are applied through the Isaac Sim/PhysX physics backend to simulate object motion and scene interactions. The extension is robot- and task-agnostic, accommodating diverse airflow configurations, target objects, and learning environments. Figure~\ref{fig:showcase} illustrates the use of the extension with various robotic platforms and target objects.

Specifically, the extension supports surface sampling for cuboids, spheres, cylinders, and arbitrary mesh-based objects, and provides several ready-to-use object models: a cube, a sphere, and a plastic bottle. During initialization, the surface of each user-specified target object is decomposed once into a set of primitives:
\begin{equation}
\mathcal{P}=\left\{\mathbf{p}_i,\mathbf{n}_i,A_i\right\},
\label{eq:surface_samples}
\end{equation}
where \(\mathbf{p}_i\) and \(\mathbf{n}_i\) denote the position and outward surface normal of the \(i\)-th primitive, respectively, and \(A_i\) is its associated surface area\footnote{All vectors in \eqref{eq:surface_samples}--\eqref{eq:airflow_wrench} are expressed in the world frame.}.

\begin{figure}[!t]
    \centerline{
        \includegraphics[width=0.9\linewidth]{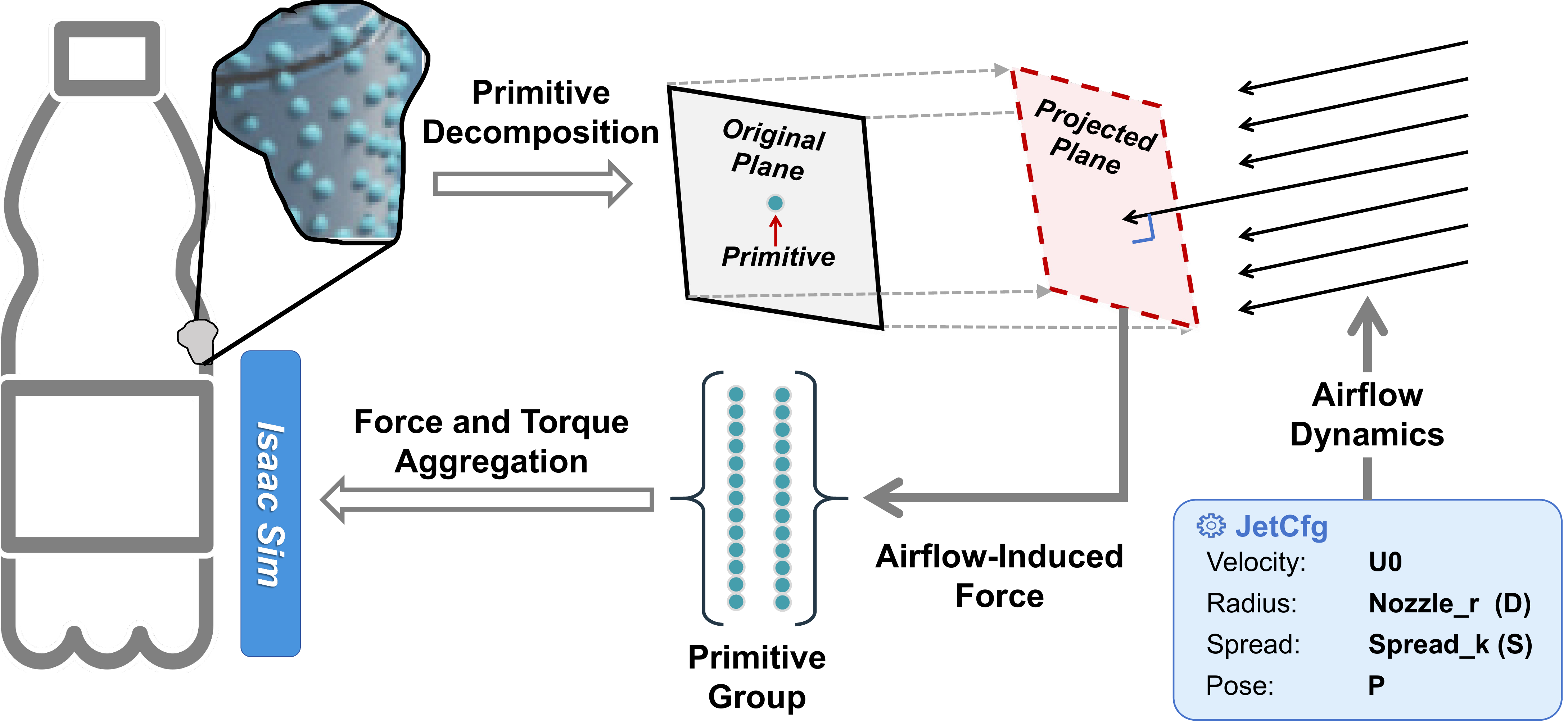}}
    \vspace{-2mm}
    \caption{Force computation pipeline of the extension. The object’s surface is decomposed into primitives, and local airflow-induced forces are computed based on the airflow configuration and each primitive’s projected area normal to the blower axis. These forces are then aggregated into the net force and torque on the object and applied in Isaac Sim.}
    \vspace{-4mm}
    \label{fig:extension}
\end{figure}

The local airflow velocity $\mathbf{u}_i$ at each primitive is computed using the airflow model in \eqref{eq:gaussian_model}. Following~\cite{shademan2020effects}, the corresponding aerodynamic force is then calculated as:
\begin{equation}
\mathbf{F}_i=\frac{1}{2}\rho C_D\left\|\mathbf{u}_i\right\|_2^2A_i\left[-\hat{\mathbf{e}}_n^{T}\mathbf{n}_i\right]_+\hat{\mathbf{e}}_n,
\label{eq:local_airflow_force}
\end{equation}
where \(\rho\) is the air density, \(C_D\) is the drag coefficient of the primitive shape, and \([\cdot]_+\) denotes the positive operator. In this equation, only primitives facing the incoming airflow contribute to the resulting force.

By integrating over all surface primitives, the force acting on the target object and the torque about its center of mass (CoM) are computed as:
\begin{equation}
\mathbf{F}=\sum_i\mathbf{F}_i,\qquad \boldsymbol{\tau}=\sum_i(\mathbf{p}_i-\mathbf{p}_o)\times\mathbf{F}_i,
\label{eq:airflow_wrench}
\end{equation}
where \(\mathbf{p}_o\) denotes the CoM of the target object.

The force and torque are applied to the rigid body through the external-force interface of Isaac Lab. Compared with a simplified airflow force evaluated only at the object center, the proposed surface-integration method captures the eccentric torque induced by object orientation and asymmetric airflow coverage. The extension also provides an interface for applying a reaction force to the rigid body link carrying the blower. The force is obtained by linearly scaling the rated thrust with the airflow action. This approximates the disturbance exerted by the high-speed blower on the physical robot and further reduces the sim-to-real gap. The computation pipeline is shown in Fig.~\ref{fig:extension}.

\section{Mechatronic Design} \label{Sec:Mechatronic}

\begin{figure}[t]
    \centerline{
    \includegraphics[width=0.9\linewidth]{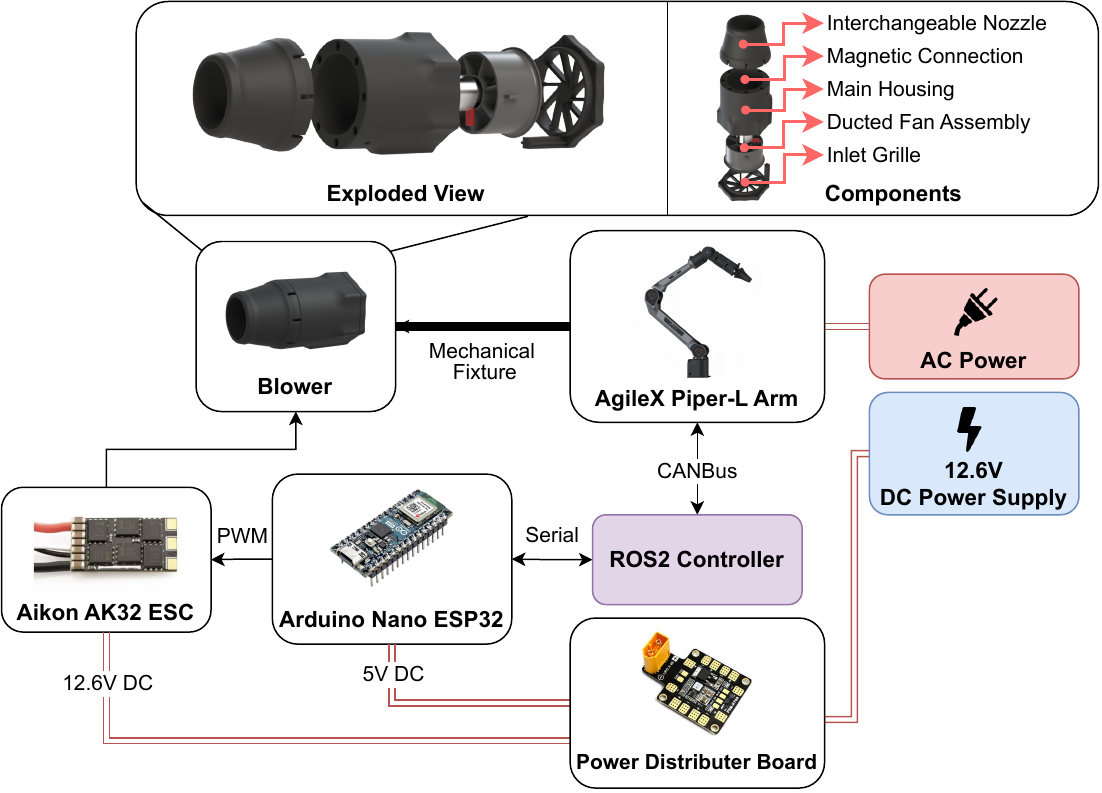}}
    \vspace{-2mm}
    \caption{Exploded view of the custom blower and system architecture showing control board interfaces and power distribution.}
    \label{fig:circuit_diagram}
    \vspace{-3mm}
\end{figure}

To enable airflow actuation on the Piper-L robotic arm from AgileX Robotics, we designed a compact blower module mounted directly below its gripper. The module houses a \(40\,\mathrm{mm}\) \emph{XFly Galaxy X4} electric ducted fan (EDF) in a 3D-printed enclosure. The EDF is driven by an \emph{Aikon AK32} electronic speed controller (ESC) and can be powered by an external supply within the nominal voltage range of 3S--4S batteries (11.1--14.8~V nominal; 12.6--16.8~V at full charge). An \emph{Arduino Nano ESP32} generates PWM signals to regulate the blower motor speed. All connections are summarized in the circuit diagram in Fig.~\ref{fig:circuit_diagram}.
Eight N52 neodymium magnets provide a detachable connection at the blower outlet to accommodate different nozzle configurations. In this work, we provide a nozzle design with an outlet diameter of \(31.7\,\mathrm{mm}\) and an outlet area corresponding to \(80\%\) of the EDF fan-swept area (FSA). The CAD files and bill of materials will be added to the website upon acceptance.

\section{Training Framework} \label{Sec:Policy}

This section presents the reinforcement learning framework for policy training, including the task formulation, training architecture, reward design, and training settings.

\subsection{Task Formulation}
To evaluate the proposed framework using a Piper-L robot arm\footnote{Compared with 7-DoF arms, the Piper-L robotic arm constrains the achievable combinations of nozzle position and orientation, limiting the range of manipulation achievable through airflow alone.}, we design two complementary tasks: a HPB task and a RR task.

In the HPB task, the policy has to move an object from a random initial position to a random 2D target region to evaluate manipulation beyond the arm's direct reach. In the RR task, the policy rotates a cuboid from a random initial orientation to the target orientation while keeping it within the predefined area, to evaluate airflow-based orientation control.

Both tasks are formulated as Markov decision processes (MDPs):
\begin{equation}
    \begin{aligned}
        &\mathcal{M}^{(J)} =
        \left(\mathcal{S}^{(J)}, \mathcal{A}, p, R^{(J)}, \gamma\right), J \in \{\mathrm{HPB}, \mathrm{RR}\},
    \end{aligned}
\end{equation}
where \(\mathcal{S}^{(J)}\) and \(\mathcal{A}\) denote the task-specific state space and the shared action space, respectively; \(p\) denotes the environment transition dynamics; \(R^{(J)}\) is the task-specific reward function; and \(\gamma\) is the discount factor. 

For each task, the policy outputs joint position commands and a normalized blower command and is optimized to maximize the expected discounted return\footnote{\([\cdot]_t\) denotes the value of the corresponding quantity of \([\cdot]\) at time \(t\).}:
\begin{equation}
    \begin{aligned}
        \max_{\theta}
        \mathbb{E}_{\pi_\theta}
        \left[
        \sum_{t=0}^{T} \gamma_t R_t^{(J)}
        \right].
    \end{aligned}
\end{equation}

Proximal policy optimization (PPO) is adopted to train the policies. The two tasks share the same training hyperparameters and optimization method, summarized in Table~\ref{tab:ppo_hyperparameters}. The actor--critic structures, reward design, and domain randomization are described in the following subsections.

\begin{table}[t]
    \caption{Key PPO hyperparameters shared by the two tasks.}
    \label{tab:ppo_hyperparameters}
    \centering
    \scriptsize
    \renewcommand{\arraystretch}{1.12}
    \begin{tabular}{l c l c}
        \hline
        \textbf{Parameter} & \textbf{Value} &
        \textbf{Parameter} & \textbf{Value} \\
        \hline
        Learning rate & $3\times10^{-4}$ &
        Clipping & 0.2 \\
        Discount factor, $\gamma$ & 0.995 &
        GAE parameter, $\lambda$ & 0.95 \\
        Epochs & 6 &
        Mini-batches & 8 \\
        Value-loss coefficient & 1.0 &
        Entropy coefficient & 0.005 \\
        Gradient norm & 1.0 &
        Initial STD & 1.0 \\
        \hline
    \end{tabular}
    \vspace{-4mm}
\end{table}

\subsection{Asymmetric Actor--Critic Architecture}

Both tasks are trained using an asymmetric actor--critic architecture\cite{pinto2017asymmetric}. Both the actor and critic are implemented as three-hidden-layer MLPs with
ELU activations and hidden dimensions of \([512,\,256,\,128]\).

The actor takes observations available during deployment as input:
\begin{equation}
\mathbf{o}^{\mathrm{actor}}_t =
\left[
\mathbf{q}^{\mathrm{hist}}_t,\,
\dot{\mathbf{q}}^{\mathrm{hist}}_t,\,
\mathbf{x}_{b,t},\,
\mathbf{\psi}_{b,t},\,
\mathbf{g}^{(J)}_t,\,
\mathbf{a}_{t-1}
\right],
\label{eq:actor_observation}
\end{equation}
where \(\mathbf{q}^{\mathrm{hist}}_t\) and
\(\dot{\mathbf{q}}^{\mathrm{hist}}_t\) denote the 10-step histories of
joint positions and velocities; \(\mathbf{x}_{b,t}\) and \(\mathbf{\psi}_{b,t}\) denote the
object position and heading; \(\mathbf{g}^{(k)}_t\) is the task goal, i.e., a
2D target position for HPB and an additional 1D orientation for RR; and \(\mathbf{a}_{t-1}\) is the previous action vector.

The critic receives additional privileged information during training:
\begin{equation}
\mathbf{o}^{\mathrm{critic}}_t =
\left[
\mathbf{o}^{\mathrm{actor}}_t,\,
\dot{\mathbf{x}}_{b,t},\,
\boldsymbol{\omega}_{b,t},\,
\Delta\mathbf{p}_{g,t}^{(J)},\,
T,\,
\boldsymbol{\eta}_{n,t}
\right],
\label{eq:critic_observation}
\end{equation}
where \(\dot{\mathbf{x}}_{b,t}\) and \(\boldsymbol{\omega}_{b,t}\) are the object's
linear and angular velocities, \(\Delta\mathbf{p}_{g,t}^{(J)}\) is the object-to-goal displacement, \(T\) denotes the target tolerance, and
\(\boldsymbol{\eta}_{n,t}\) contains the nozzle-axis direction and its
horizontal and downward components.

\begin{table}[t]
    \caption{Reward function terms and weights.}
    \label{tab:reward_terms}
    \centering
    \scriptsize
    \setlength{\tabcolsep}{3pt}
    \renewcommand{\arraystretch}{1.12}
    \begin{tabular}{p{0.31\columnwidth} p{0.52\columnwidth} c}
        \hline
        \textbf{Reward term} & \textbf{Formula} & \textbf{Weight} \\
        \hline
        \multicolumn{3}{l}{\textit{HPB Task-level Rewards}} \\
        Dense goal reward &
        \(1-\tanh(d_t/\sigma_d)\) &
        2.0 \\
        
        Goal progress &
        \((d_{t-1}-d_t)/\sigma_d\) &
        3.0 \\
        
        Goal inside &
        \(\mathbb{I}_{\mathrm{in}}\) &
        8.0 \\
        
        Goal stable &
        \(\mathbb{I}_{\mathrm{in}}
        \exp(-\|\mathbf{v}_{b,xy}\|_2^2/\sigma_v^2)\) &
        2.0 \\
        
        Nozzle--object range &
        \(\exp(-(d-d^{*})^2/\sigma_{range}^2)\) &
        0.05 \\
        
        Airflow toward goal &
        \(\max(\mathbf{v}_{b,xy}^{T}\hat{\mathbf{e}}_g,\,0)
        {u}_t g_{\mathrm{ready}}\) &
        6.0 \\

        Directed airflow force &
        \(\mathbb{I}_{\mathrm{act}}
        \operatorname{clip}(F_{g}/F_s,0,1)\) &
        1.0 \\
        \hline
        
        \multicolumn{3}{l}{\textit{RR Task-level Rewards}} \\
        Dense goal reward &
        \(1-\tanh(d_t/\sigma_d)\) &
        2.0 \\
        
        Goal progress &
        \((d_{t-1}-d_t)/\sigma_d\) &
        3.0 \\
        
        Goal inside &
        \(\mathbb{I}_{\mathrm{in}}\) &
        2.0 \\
        
        Orientation alignment &
        \(g_{\mathrm{area}}
        (1-2\delta_t/\pi)\mathbb{I}_{\mathrm{valid}}\) &
        1.0 \\
        
        Orientation progress &
        \(g_{\mathrm{area}}
        (\delta_{t-1}-\delta_t)\mathbb{I}_{\mathrm{valid}}/\sigma_\theta\) &
        2.0 \\
        
        Pose achieved &
        \(\mathbb{I}_{\mathrm{pose}}\) &
        4.0 \\
        
        Pose stable &
        \(\mathbb{I}_{\mathrm{pose}}
        \exp\!\left(
        -\|\mathbf{v}_{b,xy}\|_2^2/\sigma_v^2
        -\|\boldsymbol{\omega}_{b,z}\|_2^2/\sigma_\omega^2
        \right)\) &
        2.0 \\
        
        Nozzle--object range &
        \(\exp(-(d-d^{*})^2/\sigma_{range}^2)\) &
        0.05 \\
        
        Airflow toward goal &
        \(\max(\mathbf{v}_{b,xy}^{T}\hat{\mathbf{e}}_g,\,0)
        {u}_t g_{\mathrm{ready}}\) &
        6.0 \\
        
        Directed airflow force &
        \(\mathbb{I}_{\mathrm{act}}
        \operatorname{clip}(F_{g}/F_s,0,1)\) &
        1.0 \\
        \hline
        
        \multicolumn{3}{l}{\textit{Shared Regularization}} \\
        Airflow energy &
        \(s_{\mathrm{air}}u_t^2\) &
        $-0.08$ \\
        
        Airflow rate &
        \(s_{\mathrm{air}}(u_t-u_{t-1})^2\) &
        $-0.08$ \\
        
        Arm action deviation &
        \(\|\mathbf{q}^{\mathrm{cmd}}_t-\mathbf{q}_0\|_2^2\) &
        $-0.005$ \\
        
        Arm action rate &
        \(\|\mathbf{a}^{\mathrm{arm}}_t-
        \mathbf{a}^{\mathrm{arm}}_{t-1}\|_2^2\) &
        $-0.08$ \\
        
        Joint velocity &
        \(\|\dot{\mathbf{q}}_t\|_2^2\) &
        $-0.03$ \\

        End-effector acceleration &
        \(s_{\mathrm{ee}}\|\mathbf{a}_t^{\mathrm{ee}}\|_2^2\) &
        $-0.002$ \\
        \hline
        
        \multicolumn{3}{l}{\textit{Shared Penalty}} \\
        Airflow before ready &
        \(s_{\mathrm{air}}u_t^2
        (1-Q_{\mathrm{ready}})\) &
        $-0.40$ \\
        
        Nozzle too low &
        \(s_{\mathrm{air}}u_t d_{\mathrm{low}}^2\) &
        $-0.10$ \\
        
        Downward nozzle pitch &
        \(s_{\mathrm{air}}u_t
        \theta_{\mathrm{down}}/\theta_0\) &
        $-0.02$ \\
        
        Robot contacts &
        \(N_{\mathrm{contact}}\) &
        $-2.0$ \\
        
        Object out of workspace &
        \(\mathbb{I}[\mathbf{p}_b\notin\mathcal{W}]\) &
        $-200.0$ \\

        Object tumbling &
        \(\mathbb{I}_{\mathrm{tumble}}\) &
        $-50.0$ \\
        \hline
        \multicolumn{3}{p{\dimexpr\columnwidth-2\tabcolsep\relax}}{%
        \raggedright
        \textit{Notations:}
        \(d\): distances;
        \(\sigma,F_s,\theta_0\): normalization scales;
        \(\mathbb I\): condition indicators;
        \(\mathbf p,\mathbf v,\boldsymbol\omega\):
        position, linear/angular velocities;
        \(\hat{\mathbf e}_g\): goal direction;
        \(u\): airflow command;
        \(g\): reward gates;
        \(F_g\): goal-directed airflow force;
        \(\delta\): heading error;
        \(s\): penalty scales;
        \(\mathbf q,\dot{\mathbf q}\): joint positions/velocities;
        \(\mathbf a\):
        accelerations.
        \(Q_{\mathrm{ready}}\): nozzle readiness;
        \(\theta_{\mathrm{down}}\): downward nozzle pitch;
        \(N_{\mathrm{contact}}\): undesired contact;
        \(\mathcal W\): workspace.
        \par
    }\\
    \hline
    \end{tabular}
    \vspace{-4mm}
\end{table}

\vspace{-4mm}
\subsection{Reward Design}

The task-specific total reward is defined as:
\begin{equation}
R_\mathrm{total}^{(J)} =
R_{\mathrm{task}}^{(J)} \cdot \Delta t
+ R_{\mathrm{penalty}} \cdot \Delta t
+ R_{\mathrm{regularization}} \cdot \Delta t ,
\label{eq:total_reward}
\end{equation}
where \(R_{\mathrm{task}}^{(J)}\) denotes the task-level reward, \(R_{\mathrm{regularization}}\) denotes the regularization reward, \(R_{\mathrm{penalty}}\) denotes the penalty terms, and \(\Delta t\) is the control interval. The detailed rewards are summarized in Table~\ref{tab:reward_terms}, and the design of each component is described in the following subsections.

\begin{table}[t]
    \caption{Domain randomization parameters.}
    \label{tab:domain_randomization}
    \centering
    \scriptsize
    \renewcommand{\arraystretch}{1.15}
    \begin{tabular}{l c}
        \toprule
        \textbf{Parameter} & \textbf{Range} \\
        \midrule
        Object mass &
\([0.8,\,1.2]\times m_{\mathrm{design}}\) \\
        
        Object static friction &
        \([0.8,\,1.2]\) \\
        
        Object dynamic friction &
        \([0.7,\,1.1]\) \\
        
        Joint stiffness &
        \([0.8,\,1.2]\) \\
        
        Joint damping &
        \([0.8,\,1.2]\) \\
        
        Outlet airflow velocity, \(U_0\) &
        \([24.5,\,30.0]\) m/s \\
        
        Spreading rate, \(S\) &
        \([0.0689,\,0.0840]\) \\
        \bottomrule
    \end{tabular}
    \vspace{-2mm}
\end{table}

\begin{table}[t]
    \caption{Curriculum settings for HPB and RR.
    Step counts are expressed in thousands (k).}
    \label{tab:curriculum_settings}
    \centering
    \scriptsize
    \setlength{\tabcolsep}{2pt}
    \renewcommand{\arraystretch}{1.15}
    \begin{tabular}{@{}
        p{0.32\columnwidth}
        p{0.19\columnwidth}
        p{0.19\columnwidth}
        p{\dimexpr0.20\columnwidth-12pt\relax}
        @{}}
        \toprule
        \textbf{Curriculum} & \textbf{Initial} &
        \textbf{Final} & \textbf{Update} \\
        \midrule

        \multicolumn{4}{@{}l}{\textit{HPB Task Curriculum}} \\
        Square target side length (m) &
        \(0.24\) & \(0.20\) & Success rate \\

        Target centroid \(x\)-range (m) &
        \([0.55,0.75]\) & \([0.40,1.40]\) &
        Success rate \\

        Target centroid \(y\)-range (m) &
        \([-0.08,0.08]\) & \([-0.15,0.15]\) &
        Success rate \\

        Air penalty, \(s_{\mathrm{air}}\) &
        0 & 1 & 5--15k steps \\
        \midrule

        \multicolumn{4}{@{}l}{\textit{RR Task Curriculum}} \\
        Heading error tolerance (\(^\circ\)) &
        \(30^\circ\) & \(10^\circ\) & Success rate \\

        Success hold time (s) &
        \(0.50\) & \(1.00\) & Success rate \\

        Heading command range (\(^\circ\)) &
        \(90^\circ\pm20^\circ\) & \(90^\circ\pm45^\circ\) & Success rate \\

        Air penalty, \(s_{\mathrm{air}}\) &
        0 & 1 & 30--80k steps \\
        \bottomrule
    \end{tabular}
    \vspace{-3mm}
\end{table}

\subsubsection{HPB Task-level Reward}
This task reward guides the object toward the target region while promoting effective airflow manipulation. \textit{Dense} and \textit{progress} rewards encourage goal-directed motion, whereas \textit{inside} and \textit{stable} rewards encourage the object to enter the target region and remain stationary. The \textit{nozzle--object range} reward promotes an effective airflow distance. The \textit{airflow toward goal} reward encourages goal-directed object motion when the blower is activated and is modulated by \(g_{\mathrm{ready}}\), which evaluates nozzle placement and axis direction relative to the object motion.

\subsubsection{RR Task-level Reward}
Building on the HPB task-level rewards, this reward set further introduces dense and progress rewards for orientation alignment. Pose achievement and stability rewards encourage the object to remain stationary within the target region while maintaining the desired orientation. The nozzle--object range and airflow toward goal rewards are also retained.

\subsubsection{Shared Regularization and Penalties}
The regularization terms promote smooth and efficient policy execution. The airflow energy and rate terms penalize excessive airflow intensity and rapid changes in airflow, respectively, while the action-rate and joint-velocity terms constrain abrupt changes in the arm control policy and aggressive joint motions. Once the object enters the target region, the arm action deviation term further encourages the robot arm to return to its default posture.

The penalty terms discourage airflow activation when the nozzle is not correctly positioned, too low, or at a large downward angle. Penalties also apply to undesired robot contacts, object motion outside the workspace, and object tumbling.

\subsubsection{Domain Randomization, Termination, and Curriculum}

To reduce the sim-to-real gap arising from modeling and control uncertainties, the following parameters are randomized during training: object physical properties, robot joints' impedance, and identified airflow parameters, as summarized in
Table~\ref{tab:domain_randomization}\footnote{The randomization ranges for $U_0$ and $S$ are set to $\pm 10\%$ of the corresponding identified values for bare-EDF in Sec.~\ref{Sec:Experiment}-A.}. 

An episode terminates upon timeout, object being moved outside the workspace\footnote{The workspace is defined as
\(x\in[0.05,1.65]\)~m, \(y\in[-0.40,0.40]\)~m.}, excessively low end-effector height, or collision between the blower and the object. For the RR task, we also add gripper-object contact as a termination condition.

Table~\ref{tab:curriculum_settings} summarizes the curriculum used in both HPB and RR policy training. Task difficulty and penalty scales are gradually increased to facilitate learning.

\vspace{-3mm}
\section{Experimental Results} \label{Sec:Experiment}
This section presents airflow identification and validation experiments, as well as real-world experiments to evaluate the trained policies. Further details of the experimental settings and results are shown in the accompanying video.
\vspace{-4mm}
\subsection{Airflow Identification}

\subsubsection{Numerical and Experimental Setup}
To establish a comparison baseline, we perform a Computational Fluid Dynamics (CFD) simulation of an ideal turbulent round jet in OpenFOAM~v14~\cite{openfoam}, solving the incompressible RANS equations with the standard $k$-$\epsilon$ model and SIMPLE algorithm \cite{versteeg1995introduction}. A simulated axisymmetric $5^\circ$ wedge (a thin angular slice around the jet axis) extends $50D$ axially and $15D$ radially, with a $4{:}1$ graded $200\times60\times1$ mesh of $12{,}000$ cells. The inlet uses a top-hat profile with $U_{0, \mathrm{EDF}}=27.28~\mathrm{m/s}$ and $U_{0, \mathrm{nozzle}}=30.5~\mathrm{m/s}$, $5\%$ turbulence intensity, and $1.516\times10^{-5}~\mathrm{m^2/s}$ kinematic viscosity. After 500 iterations, all initial residuals are below $10^{-4}$. The model excludes motor-hub blockage and blade-induced swirl.

Real-world airflow measurements were recorded at centerline and radial locations using a \emph{HoldPeak HP-5000Y} anemometer ($\pm5\%$ accuracy). The EDF is powered by $12.6~\mathrm{V}$ and set to $50\%$ PWM, drawing approximately $1.89~\mathrm{A}$. Three readings were taken per location after $5~\mathrm{s}$ settling period for steady airflow conditions to develop. Centerline measurements were obtained by translating the aligned blower along its axis, while radial measurements were taken in both directions at $x/D=5,10,$ and $15$. The reference diameters were $D_\mathrm{EDF}=40~\mathrm{mm}$ and $D_\mathrm{nozzle}=31.7~\mathrm{mm}$.

\begin{table*}[!t]
    \centering
    \caption{Centerline model-identification results for CFD simulations and experimental EDF measurements,\\ including fitted parameters and their standard errors (SE).}
    \label{tab:centerline_results}
    \begin{tabular}{llcccccc}
        \toprule
        Dataset & Model & RMSE ($\mathrm{m/s}$) & $R^2$ & MARE (\%) & $k \pm \mathrm{SE}$ ($\mathrm{m}^{-1}$) & $B \pm \mathrm{SE}$ ($\mathrm{-}$) & $x_0 \pm \mathrm{SE}$ ($\mathrm{mm}$) \\
        \midrule
        CFD ($D=40~\mathrm{mm}$) & Exponential & 1.74 & 0.927 & 30.2 & 2.293 $\pm$ 0.044 & --- & --- \\
        & \textbf{Inverse} & \textbf{0.35} & \textbf{0.997} & \textbf{3.4} & --- & 3.324 $\pm$ 0.037 & -36.1 $\pm$ 2.2 \\
        CFD ($D=31.7~\mathrm{mm}$) & Exponential & 1.94 & 0.927 & 30.2 & 2.893 $\pm$ 0.055 & --- & --- \\
        & \textbf{Inverse} & \textbf{0.39} & \textbf{0.997} & \textbf{3.4} & --- & 5.324 $\pm$ 0.037 & -28.6 $\pm$ 1.7 \\
        EDF without Nozzle & \textbf{Exponential} & \textbf{0.61} & \textbf{0.994} & \textbf{6.6} & 2.648 $\pm$ 0.065 & --- & --- \\
        & Inverse & 1.07 & 0.981 & 10.9 & --- & 4.573$ \pm$ 0.279 & 15.9 $\pm$ 14.1 \\
        EDF with $0.8$-FSA Nozzle & Exponential & 1.00 & 0.988 & 8.7 & 3.170 $\pm$ 0.077 & --- & --- \\
        & \textbf{Inverse} & \textbf{0.90} & \textbf{0.991} & \textbf{7.2} & --- & 4.565 $\pm$ 0.198 & 29.9 $\pm$ 9.7 \\
        \bottomrule
    \end{tabular}
    \vspace{-5mm}
\end{table*}

\subsubsection{Centerline Model Identification}

\begin{figure}[t]
    \centering
    \includegraphics[width=0.95\linewidth]{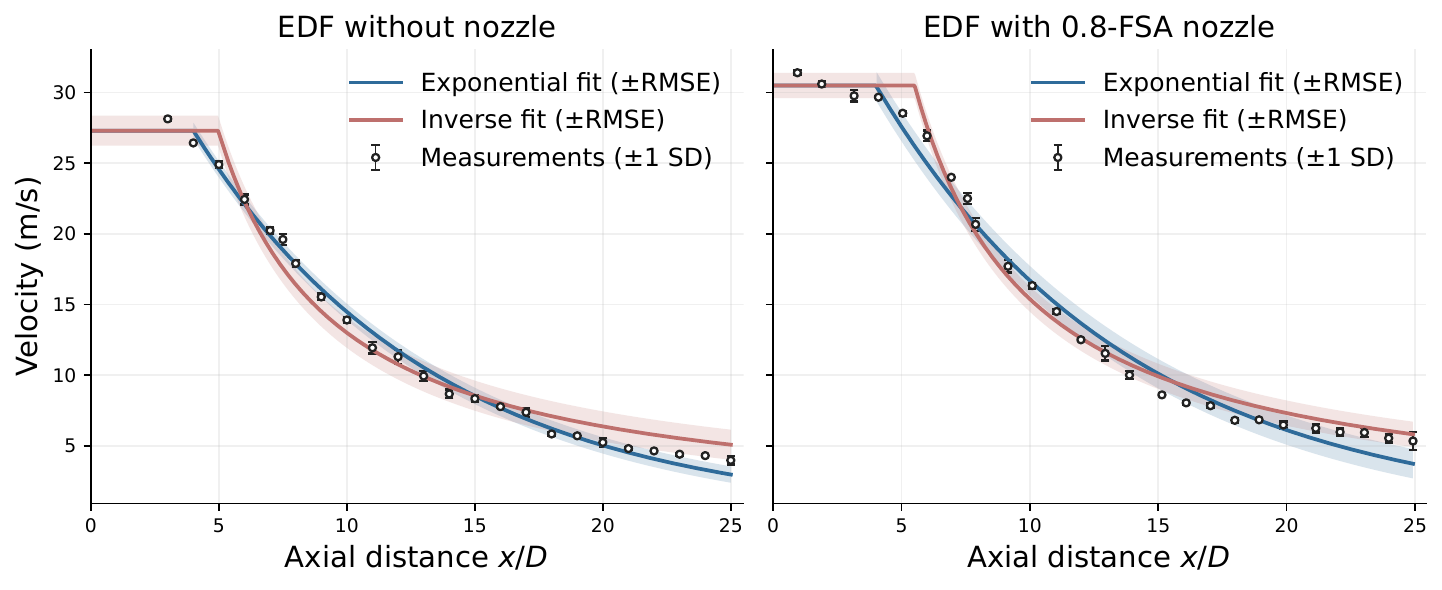}
    \vspace{-2mm}
    \caption{Centerline velocity comparison for the no-nozzle and nozzle configurations. Axial distance is normalized by $D$.}
    \label{fig:centerline_comparison}
    \vspace{-3mm}
\end{figure}

Both centerline models (See Sec.~\ref{Sec:Modeling}, \eqref{eq:exp_model}, \eqref{eq:inv_model}) were evaluated against the CFD simulation and experimental data at identical locations. The exponential model used $x_c=4D$, while the shifted-inverse model was capped at $U_0$ near the outlet. Table~\ref{tab:centerline_results} summarizes the errors for the CFD, EDF without nozzle, and with nozzle cases, while Fig.~\ref{fig:centerline_comparison} compares the normalized measurements and fitted models.

Against CFD, the shifted-inverse model reduced the RMSE by $79.81\%$ relative to the exponential model for both blower configurations, as expected for a developed turbulent round jet. For the EDF without a nozzle, however, the exponential model achieved a lower RMSE of $0.61~\mathrm{m/s}$ versus $1.07~\mathrm{m/s}$. 

The $0.8$-FSA nozzle produced a more axisymmetric round-jet-like flow, for which the shifted-inverse model achieved a lower RMSE of $0.90~\mathrm{m/s}$ versus $1.00~\mathrm{m/s}$. The maximum experimental RMSE and mean absolute error were $1.07~\mathrm{m/s}$ and $2.22~\mathrm{m/s}$, respectively, partly reflecting the anemometer accuracy. Ultimately, the results favor the shifted-inverse model since it performs better overall, even though the exponential model remains a competitive approximation when it comes to capturing the centerline velocity decay of a free turbulent round jet. 

\subsubsection{Radial Profile Identification}

\begin{figure}[t]
    \centering
    \includegraphics[width=0.95\linewidth]{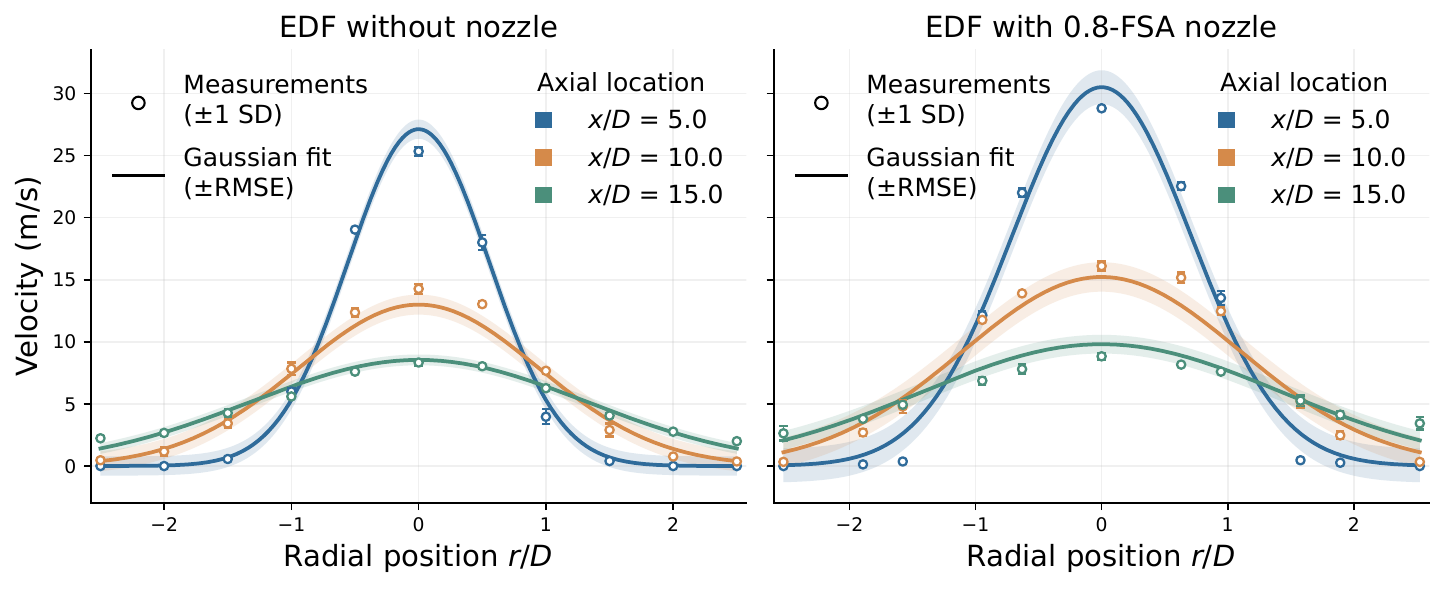}
    \vspace{-2mm}
    \caption{Measured and fitted radial velocity profiles for no-nozzle and nozzle configurations. Radius is normalized by $D$, and color identifies the axial station $x/D$.}
    \label{fig:radial_comparison}
    \vspace{-5mm}
\end{figure}

\begin{figure*}[!t]
    \centering
    \vspace{-3mm}
    \includegraphics[width=0.85\textwidth]{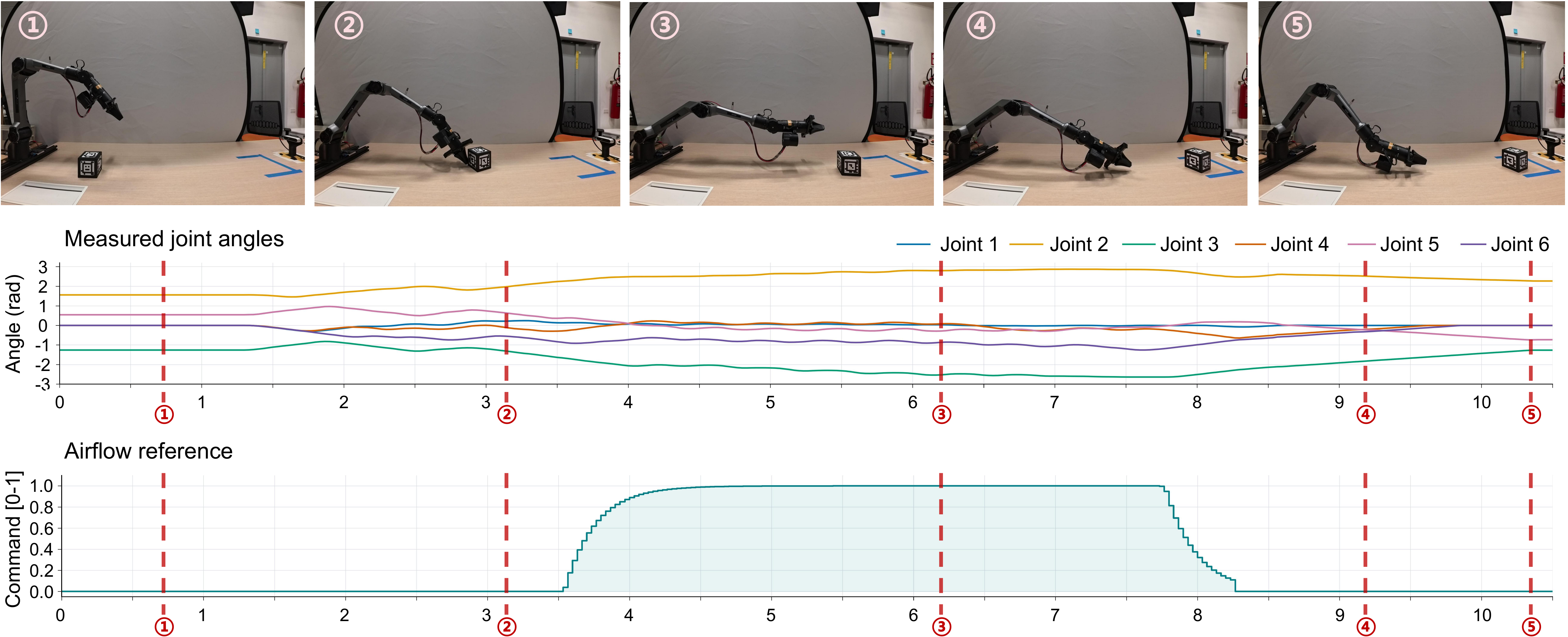}\
    \vspace{-3mm}
    \caption{
    HPB task policy deployment. The arm first pushes the cube into a position suitable for blowing (\ding{172}–\ding{173}). It then aims the blower at the cube, activates the airflow, and adjusts the blowing direction to guide the cube toward the target region (\ding{174}). Once the cube enters the target region, the airflow command decreases to zero, leaving the cube in the target region (\ding{175}–\ding{176}). The plots below show the measured angles of the six joints and the airflow reference; the red dashed lines mark the times corresponding to the images above.}
    \label{fig:HPB}

    \vspace{0.2cm}

    \includegraphics[width=0.85\textwidth]{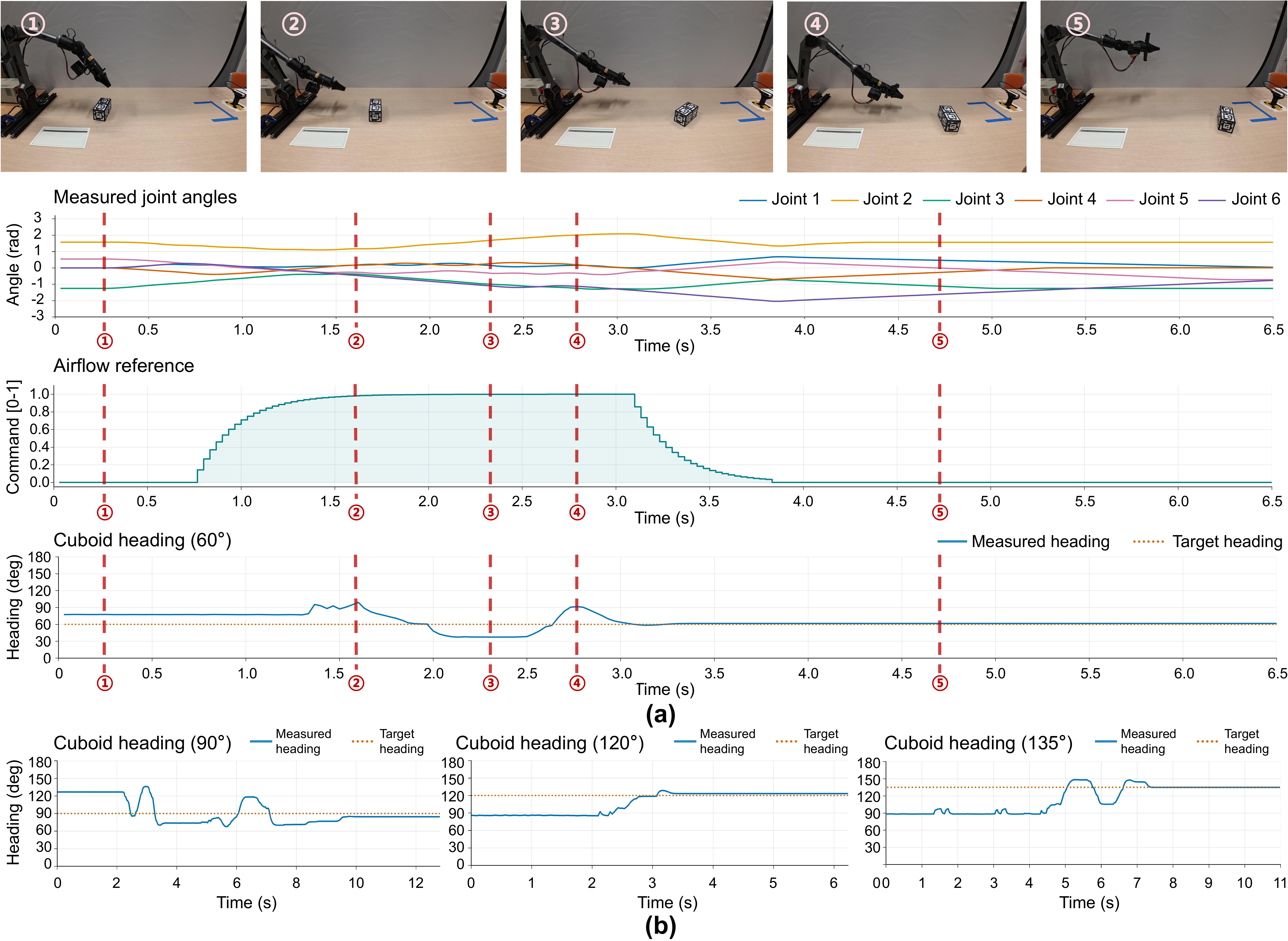}
    \vspace{-3mm}
    \caption{RR task policy deployment. (a) For a target heading of \(60^\circ\), the arm continuously adjusts the blowing direction to rotate the cuboid using contact-free airflow (\ding{172}--\ding{175}). Once the cuboid reaches the target heading, the airflow command gradually decreases to zero (\ding{176}). The plots below show the measured angles of the six joints, the airflow reference, and the cuboid heading; the red dashed lines mark the times corresponding to the images above. (b) Measured and target headings for target angles of \(90^\circ\), \(120^\circ\), and \(135^\circ\).}
    \label{fig:RR}
    \vspace{-5mm}
\end{figure*}

Radial measurements at three centerline stations were fitted with Gaussian profiles. To decouple them from the centerline models, both the amplitude and $\sigma_r$ were fitted directly to the experimental data. Opposite radial measurements differed by only $3\%$--$7\%$, indicating approximate axisymmetry within the anemometer accuracy. The fits achieved $R^2\geq0.960$ and $R^2\geq0.970$ for the EDF with and without nozzle configurations, respectively. As shown in Fig.~\ref{fig:radial_comparison}, $\sigma_r$ increased approximately linearly with distance. The CFD, EDF without nozzle, and EDF with nozzle spreading rates were $0.096$, $0.077$, and $0.083$, corresponding to width angles of $5.50^\circ$, $4.38^\circ$, and $4.73^\circ$, respectively.

\vspace{-3mm}
\subsection{Policy Deployment}
\subsubsection{Policy Training}
Training was conducted on a computer equipped with an NVIDIA RTX 5090 GPU and an Intel U7 265K CPU, using 4096 parallel environments for 5000 iterations, with a total training time of 5.7 h (HPB) and 6.6 h (RR). In simulation, the trained policies achieved success rates of 88\% on the HPB task and 92\% on the RR task.

\subsubsection{Experimental Setup}
For real-world experiments, a calibrated RealSense D435f camera was mounted on the table to detect objects, while AprilTags~\cite{wang2016apriltag} attached to the objects enabled real-time orientation estimation.

\subsubsection{Real-world Experiment Results}
\paragraph{HPB Task}

In the HPB task, when the target region is set within the gripper's reach, the policy primarily uses contact-based manipulation, consistent with the airflow-use penalty. When the target region is beyond the arm's direct reach, the policy first repositions the cube by pushing and then uses airflow to blow it into the target region (see Fig.~\ref{fig:HPB}).

To evaluate the policy’s success rate over repeated trials, we defined a fixed square target region with a width of 20 cm, centered 1.3 m in front of the robot base, and conducted 20 trials. The cube successfully entered the target region in 16 trials, achieving a success rate of 80\%. 

Across the repeated trials, failures fell into two main categories: overly aggressive motions during pushing that displaced the cube outside regions where airflow or pushing could correct it; and arm oscillations during blowing that caused unintended contact with the object, pushing it in undesired directions. Both failures are closely related to the policy’s performance during direct contact interactions and the limited margins available for recovery under the arm’s kinematic constraints.

\paragraph{RR Task}
We conducted experiments with heading targets of \(60^\circ\), \(90^\circ\), \(120^\circ\), and \(135^\circ\). The policy manipulated the cuboid using airflow only, with the arm adjusting the blowing direction to rotate the cuboid toward the target heading. Fig.~\ref{fig:RR}(a) shows the deployment detail for the \(60^\circ\) trial, while Fig.~\ref{fig:RR}(b) shows the heading trajectories for the other trials. The final headings were \(61.6^\circ\), \(84.5^\circ\), \(123.2^\circ\), and \(134.9^\circ\), all within the \(\pm10^\circ\) tolerance and a mean absolute error of \(2.6^\circ\).

\vspace{-3mm}

\section{Conclusion}\label{Sec:Conclusion}

This work developed an Isaac Lab airflow-interaction extension for massively parallel RL training, supported by experimentally identified airflow decay and spreading models and a lightweight, high-speed blower mounted alongside a robotic arm’s gripper. Building on this extension, we proposed training frameworks for HPB and RR, enabling hybrid contact-based pushing and non-contact blowing in HPB and airflow-only object reorientation in RR. In zero-shot hardware deployment, HPB achieved an 80\% success rate over 20 trials, while RR achieved a mean absolute heading error of \(2.6^\circ\) across four target headings. The arm’s kinematic constraints restrict the positions and directions from which airflow can be applied, motivating our future work on a mobile base to extend the operating range, alongside evaluation on objects with varied shapes and masses across more complex scenarios and tasks.

\vspace{-2mm}

\bibliographystyle{IEEEtran}
\bibliography{bibliography}

@article{hussein1994velocity,
  title={Velocity measurements in a high-Reynolds-number, momentum-conserving, axisymmetric, turbulent jet},
  author={Hussein, Hussein J and Capp, Steven P and George, William K},
  journal={Journal of Fluid Mechanics},
  volume={258},
  pages={31--75},
  year={1994},
  publisher={Cambridge University Press}
}

@article{albertson1950diffusion,
  title={Diffusion of submerged jets},
  author={Albertson, ML and Dai, YeB and Jensen, Ro A and Rouse, Hunter},
  journal={Transactions of the American Society of Civil Engineers},
  volume={115},
  number={1},
  pages={639--664},
  year={1950},
  publisher={American Society of Civil Engineers}
}

@article{kopitca2025remote,
  title={Remote Manipulation of Multiple Objects With Airflow Field Using Model-Based Learning Control},
  author={Kopitca, Artur and Haeri, Shahriar and Zhou, Quan},
  journal={IEEE/ASME Transactions on Mechatronics},
  year={2025},

}

@article{wu2022learning,
  title={Learning pneumatic non-prehensile manipulation with a mobile blower},
  author={Wu, Jimmy and Sun, Xingyuan and Zeng, Andy and Song, Shuran and Rusinkiewicz, Szymon and Funkhouser, Thomas},
  journal={IEEE Robotics and Automation Letters},
  volume={7},
  number={3},
  pages={8471--8478},
  year={2022},

}

@article{ball2012flow,
  title={The flow field in turbulent round free jets},
  author={Ball, CG and Fellouah, H and Pollard, A},
  journal={Progress in Aerospace Sciences},
  volume={50},
  pages={1--26},
  year={2012},
  publisher={Elsevier}
}

@inproceedings{turrisi2026binwalker,
  title={BinWalker: Development and Field Evaluation of a Quadruped Manipulator Platform for Sustainable Litter Collection},
  author={Turrisi, Giulio and Bratta, Angelo and Minelli, Giovanni and Abati, Gabriel Fischer and Rad, Amir H and Soares, Jo{\~a}o Carlos Virgolino and Semini, Claudio},
  booktitle={2026 IEEE/ASME International Conference on Advanced Intelligent Mechatronics (AIM)},
  pages={1--8},
}

@inproceedings{spahn2024demonstrating,
  title={Demonstrating Adaptive Mobile Manipulation in Retail Environments.},
  author={Spahn, Max and Pezzato, Corrado and Salmi, Chadi and Dekker, Rick and Wang, Cong and Pek, Christian and Kober, Jens and Alonso-Mora, Javier and Corbato, Carlos Hernandez and Wisse, Martijn},
  booktitle={Robotics: Science and Systems},
  year={2024}
}

@article{rossini2026concert,
  title={Concert: a modular reconfigurable robot for construction},
  author={Rossini, Luca and Romiti, Edoardo and Laurenzi, Arturo and Ruscelli, Francesco and Ruzzon, Marco and Covizzi, Luca and Baccelliere, Lorenzo and Carrozzo, Stefano and Terzer, Michael and Magri, Marco and others},
  journal={Journal of Field Robotics},
  volume={43},
  number={3},
  pages={1332--1362},
  year={2026},
  publisher={Wiley Online Library}
}

@article{ren2025object,
  title={Object-centric kinodynamic planning for nonprehensile robot rearrangement manipulation},
  author={Ren, Kejia and Wang, Gaotian and Morgan, Andrew S and Kavraki, Lydia E and Hang, Kaiyu},
  journal={IEEE Transactions on Robotics},
  year={2025},

}

@article{al2022non,
  title={A non-contact manipulation for robotic applications: a review on acoustic levitation},
  author={Al-Nuaimi, Ibrahim Ismael Ibrahim and Mahyuddin, Muhammad Nasiruddin and Bachache, Nasseer K},
  journal={IEEE Access},
  volume={10},
  pages={120823--120837},
  year={2022},

}

@inproceedings{lee2022development,
  title={Development of a leaf-sweeping robot},
  author={Lee, Wei-chen and Chiang, Wei-hsun},
  booktitle={2022 IEEE 5th International Conference on Industrial Cyber-Physical Systems (ICPS)},
  pages={1--5},
}

@inproceedings{natsu2023development,
  title={Development of an autonomous blower robot for cleaning up and collecting fallen leaves},
  author={Natsu, Nakagawa and others},
  booktitle={2023 IEEE 19th International Conference on Automation Science and Engineering (CASE)},
  pages={1--6},
}

@article{haeri2024meter,
  title={Meter-Scale Distance Manipulation of Diverse Objects with Jet-Induced Airflow Field},
  author={Haeri, Shahriar and Kopitca, Artur and Kandemir, Hakan and Zhou, Quan},
  journal={Advanced Intelligent Systems},
  volume={6},
  number={11},
  pages={2400174},
  year={2024},
  publisher={Wiley Online Library}
}

@article{xu2022dextairity,
  title={Dextairity: Deformable manipulation can be a breeze},
  author={Xu, Zhenjia and Chi, Cheng and Burchfiel, Benjamin and Cousineau, Eric and Feng, Siyuan and Song, Shuran},
  journal={arXiv preprint arXiv:2203.01197},
  year={2022}
}

@article{moon2006distributed,
  title={Distributed manipulation of flat objects with two airflow sinks},
  author={Moon, Hyungpil and Luntz, Jonathan},
  journal={IEEE Transactions on Robotics},
  volume={22},
  number={6},
  pages={1189--1201},
  year={2006},

}

@inproceedings{kopitca20263d,
  title={3D manipulation of airborne objects using patterned airflow},
  author={Kopitca, Artur and Haeri, Shahriar and Iqbal, Saad and Zang, Xiaoya and Zeng, Hao and Zhou, Quan},
  booktitle={2026 International Conference on Manipulation, Automation and Robotics at Small Scales (MARSS)},
  pages={1--6},
  organization={IEEE}
}

@article{bao2026development,
  title={Development of a Bernoulli Suction-Type Non-Contact Manipulator Capable of Controlling Object Position and Orientation During Holding},
  author={Bao, Yuanhao and Ambe, Yuichi and Takaki, Takeshi},
  journal={IEEE Robotics and Automation Letters},
  year={2026},

}

@article{jia2025non,
  title={Non-contact dexterous micromanipulation with multiple optoelectronic robots},
  author={Jia, Yongyi and Miao, Shu and Wang, Ao and Ni, Caiding and Feng, Lin and Wang, Xiaowo and Li, Xiang},
  journal={IEEE Robotics and Automation Letters},
  volume={10},
  number={5},
  pages={4412--4419},
  year={2025},

}

@article{shademan2020effects,
  title={Effects of aspect ratio and inclination angle on aerodynamic loads of a flat plate},
  author={Shademan, Mehrdad and Naghib-Lahouti, Arash},
  journal={Advances in Aerodynamics},
  volume={2},
  number={1},
  pages={14},
  year={2020},
  publisher={Springer}
}

@inproceedings{wang2016apriltag,
  title={AprilTag 2: Efficient and robust fiducial detection},
  author={Wang, John and Olson, Edwin},
  booktitle={2016 IEEE/RSJ International Conference on Intelligent Robots and Systems (IROS)},
  pages={4193--4198},
}

@misc{openfoam,
  title={{OpenFOAM}},
  url={https://openfoam.org/},
  author={{The OpenFOAM Foundation}},
}

@article{basset2022entrainment,
  title={Entrainment, diffusion and effective compressibility in a self-similar turbulent jet},
  author={Basset, Thomas and Viggiano, Bianca and Barois, Thomas and Gibert, Mathieu and Mordant, Nicolas and Cal, Ra{\'u}l Bayo{\'a}n and Volk, Romain and Bourgoin, Micka{\"e}l},
  journal={Journal of Fluid Mechanics},
  volume={947},
  pages={A29},
  year={2022},
  publisher={Cambridge University Press}
}

@misc{nvidia_isaaclab,
  author       = {{NVIDIA}},
  title        = {{Isaac Lab}},
  howpublished = {\url{https://developer.nvidia.com/isaac/lab}},
  note         = {Accessed: Oct. 16, 2026}
}

@article{versteeg1995introduction,
  title={An introduction to computational fluid dynamics},
  author={Versteeg, Henk Kaarle and Malalasekera, Weeratunge},
  year={1995},
  publisher={Longman New York}
}

@article{abramovich1984theory,
  title={The theory of turbulent jets},
  author={Abramovich, Genrikh Naumovich and Girshovich, TA and Krasheninnikov, S Iu and Sekundov, AN and Smirnova, IP},
  journal={Moscow Izdatel Nauka},
  year={1984}
}

@inproceedings{pinto2017asymmetric,
  title={Asymmetric Actor Critic for Image-Based Robot Learning},
  author={Pinto, Lerrel and Andrychowicz, Marcin and Welinder, Peter and Zaremba, Wojciech and Abbeel, Pieter},
  booktitle={Robotics: Science and Systems},
  year={2018},
  
}

@article{bai2026robotic,
  title={Robotic Non-Contact 3-D Micromanipulation by Acoustohydrodynamic Effects},
  author={Bai, Chenhao and Chen, Zhuo and Huang, Qiang and Arai, Tatsuo and Liu, Xiaoming},
  journal={IEEE Transactions on Automation Science and Engineering},
  volume={23},
  pages={7628--7641},
  year={2026},

}

@article{stewart2025non,
  title={Non-Contact Manipulation of Induced Magnetic Dipoles},
  author={Stewart, Seth and Pawelski, Joseph and Ward, Steve and Petruska, Andrew J},
  journal={IEEE Transactions on Magnetics},
  year={2025},

}

\newpage

\vfill

\end{document}